\documentclass{article}

\usepackage[final]{neurips_2025}

\usepackage[utf8]{inputenc} % allow utf-8 input
\usepackage[T1]{fontenc}    % use 8-bit T1 fonts
\usepackage{hyperref}       % hyperlinks
\usepackage{url}            % simple URL typesetting
\usepackage{booktabs}       % professional-quality tables
\usepackage{amsfonts}       % blackboard math symbols
\usepackage{nicefrac}       % compact symbols for 1/2, etc.
\usepackage{microtype}      % microtypography
\usepackage{xcolor}         % colors
\usepackage{amsmath}
\usepackage{subfig}
\usepackage{graphicx}

\makeatletter
\newcommand{\@trackname}{}
\makeatother

\title{Bridging Embodiment Gaps: Deploying Vision-Language-Action Models on Soft Robots}

\author{%
  Haochen Su\\
  EPFL\\
  Lausanne, Switzerland \\
  \texttt{haochen.su@alumni.epfl.ch}\\
  \And
  Cristian Meo \\
  LatentWorlds AI \\
  TUDelft \\
  Delft, Netherlands \\
\texttt{cristianmeo@latentworlds.ai} \\
  \AND
  Francesco Stella \\
  Embodied AI SA \\
  EPFL \\
  Lausanne, Switzerland \\
  \texttt{f.stella@embodiedai.ch}\\
  \And
  Andrea  Peirone\\
  Embodied AI SA \\
  EPFL \\
  Lausanne, Switzerland \\
  \texttt{andree.peirone@gmail.com}\\
  \And
  Kai Junge \\
  Embodied AI SA \\
  EPFL \\
  Lausanne, Switzerland \\
  \texttt{k.junge@embodiedai.ch}\\
  \And
  Josie Hughes \\
  EPFL\\
  Lausanne, Switzerland \\
  \texttt{josie.hughes@epfl.ch}\\
}

\begin{document}

\maketitle

%%%% OLD ABSTRACT
% Robotic systems are increasingly expected to operate in human-centered, unstructured environments where safety, adaptability, and generalization are essential. 
% While Vision-Language-Action (VLA) models have shown strong performance on rigid manipulators, their applicability to compliant platforms remains largely unexplored. 
% This work investigates whether VLA policies can be effectively transferred to soft continuum robots, whose deformable structures ensure safe interaction but introduce nonlinear, underactuated dynamics. 
% We present a structured finetuning and deployment pipeline on a custom soft robotic arm, evaluating two state-of-the-art VLA models (OpenVLA-OFT and $\pi_0$) across representative manipulation tasks. 
% Out-of-the-box policies fail due to embodiment mismatch, but with targeted finetuning, both models achieve high success rates on the soft robot—comparable to their performance on rigid manipulators. 
% Our findings highlight the necessity of finetuning for bridging embodiment gaps, and demonstrate that coupling VLA models with soft robots enables safe and flexible embodied AI in human-shared environments.

\begin{abstract}
% High level statment
Robotic systems are increasingly expected to operate in human-centered, unstructured environments where safety, adaptability, and generalization are essential. 
% VLAs are magic but safety issue 
Vision-Language-Action (VLA) models have been proposed as a language guided generalized control framework for real robots. However, their deployment has been limited to conventional serial link manipulators. 
% Why VLA is limited
Coupled by their rigidity and unpredictability of learning based control, the ability to safely interact with the environment is missing yet critical.
% In this work
In this work, we present the deployment of a VLA model on a soft continuum manipulator to demonstrate autonomous safe human-robot interaction. 
% More detail
We present a structured finetuning and deployment pipeline evaluating two state-of-the-art VLA models (OpenVLA-OFT and $\pi_0$) across representative manipulation tasks, and show while out-of-the-box policies fail due to embodiment mismatch, through targeted finetuning the soft robot performs equally to the rigid counterpart.
% Conclusion
Our findings highlight the necessity of finetuning for bridging embodiment gaps, and demonstrate that coupling VLA models with soft robots enables safe and flexible embodied AI in human-shared environments.
\end{abstract}

\section{Introduction}

To deploy robots in human-centric, real-world settings, they must interpret human instructions, perceive dynamic environments, and execute robust actions. Vision-Language-Action (VLA) models unify perception, language understanding, and control within a single multimodal policy\cite{survey}, offering a promising approach to these challenges. Encompassing CLIPort\cite{CLIPort}, SayCan\cite{saycan}, RT-2\cite{RT-2} and OpenVLA\cite{openvla}, VLA models have progressively improved generalization across tasks and settings. Yet, nearly all existing models and deployment focuses on rigid robotic arms, where predictable kinematics simplify control but limit safety and adaptability in human-centered environments.

Soft robots incorporate compliant or soft structures into their bodies, such that they deform in response to interactions with the environment. This makes them well suited for operating around humans as they provide intrinsic safety to the environment, are resilient to collisions, and can be robustness to environmentally uncertainty\cite{rus2015design, stella2023science}. Soft continuum manipulators, in particular, bring these benefits to manipulation\cite{dou2021soft, chen2022review}.  Currently soft arms rely on controllers that account for their underlying non-linear properties and redundancy within the structures. Deploying VLA models on such platforms remains unexplored: existing datasets and benchmarks overwhelmingly rely on rigid, serial-linked robots\cite{open_x_embodiment_rt_x_2023}, leaving open questions about embodiment transfer.

This gap poses two key challenges. First, reliance on rigid embodiments restricts VLA applicability to domains where compliance is crucial. Second, the nonlinear, underactuated dynamics of soft robots raise doubts about whether policies trained on rigid arms can generalize effectively. Addressing this challenge is critical to deploying VLAs models on soft robot arms, combining their physical safety with the human-relevant capabilities of VLAs. 

In this work, we take a step toward bridging this gap. We propose and implement a finetuning pipeline for deploying VLA models on a custom soft continuum robot, evaluating both OpenVLA-OFT\cite{openvla-oft} and $\pi_0$\cite{pi0}. Our study systematically benchmarks embodiment transfer across rigid and soft robots and compares the relative strengths of two state-of-the-art VLA models. Concretely, our contributions are:
\begin{enumerate}
    \item We introduce the first \textbf{open-source dataset of soft robot demonstrations}, enabling reproducible research on compliant embodiments.  
    \item We \textbf{benchmark OpenVLA-OFT on both rigid (UR5) and soft robots}, showing that finetuning closes the rigid-to-soft domain gap and yields comparable task success rates.  
    \item We \textbf{compare OpenVLA-OFT and $\boldsymbol{\pi_0}$ on the soft robot}: while $\pi_0$ demonstrates stronger generalization on rigid embodiments, OpenVLA-OFT achieves superior performance on the compliant platform after finetuning.  
\end{enumerate}

\section{Related Work}
Vision-Language-Action (VLA) models unify perception, language, and control for robotic agents. Early approaches such as CLIPort~\cite{CLIPort} and SayCan~\cite{saycan} demonstrated the potential of pretrained vision-language models, while large-scale efforts like RT-1~\cite{RT-1} and RT-2~\cite{RT-2} improved task coverage on rigid manipulators. More recent methods focus on temporal reasoning and efficiency, including $\pi_0$~\cite{pi0} with flow-based policies and OpenVLA-OFT~\cite{openvla-oft} with parallel decoding and continuous outputs. 

VLA models have also shown transfer between different rigid embodiments~\cite{open_x_embodiment_rt_x_2023}, suggesting a degree of generality. However, rigid robots share similar inverse kinematics and appearance, making transfer comparatively easier. In contrast, soft continuum manipulators exhibit nonlinear, underactuated dynamics and different morphology, a setting that remains unaddressed in prior VLA benchmarks\cite{libero}\cite{aloha}. Our work provides the first systematic evaluation of VLA models on a soft robotic arm. For more details about recent VLA methods, and state-of-the-art soft continuum robots' control, refer to Appendix \ref{appendix:related_work}.

\section{Methodology}
To investigate the deployment of VLA models on soft robotic systems, we adopt a structured pipeline spanning task design, data collection, preprocessing, model adaptation, and evaluation. We begin by defining three representative manipulation tasks tailored to the soft robot’s capabilities. Next, we set up a data-capturing environment to record multimodal demonstrations, which are then converted into standardized formats. Using these processed datasets, we finetune both models under comparable conditions. Finally, we perform inference and evaluate policy performance on the designed tasks, assessing both success rate and qualitative behavior.

\subsection{Robot platforms}
% Traditional collaborative robots are fundamentally built upon rigid-link architectures, utilizing discrete joints, metal components, and force/torque sensors to safely interact with human operators. While these systems have been demonstrated to be effective in structured environments and well-defined tasks, they are inherently limited by their rigidity, restricted adaptability, and need for elaborate safety measures to prevent injury in unplanned contact scenarios. 
% The soft continuum manipulator under development is a radical departure from this paradigm, employing an architectural design. Recent advancements in architectural materials applied to soft continuum manipulator \cite{helix} have provided the foundation for a novel approach that leverages geometric tunability. This novel approach enables independent control over axial and bending stiffness, thereby facilitating the integration of large, dexterous workspaces with compliant and safe physical interactions. 
% In particular, the architectural structure we design is highly stiff in the axial direction, with tunable stiffness for the bending direction. This is achieved in order to increase the payload capacity of the manipulator, due to the incompressible nature of the structure. 

\begin{figure}[h]
    \centering
    \includegraphics[width=0.9\columnwidth]{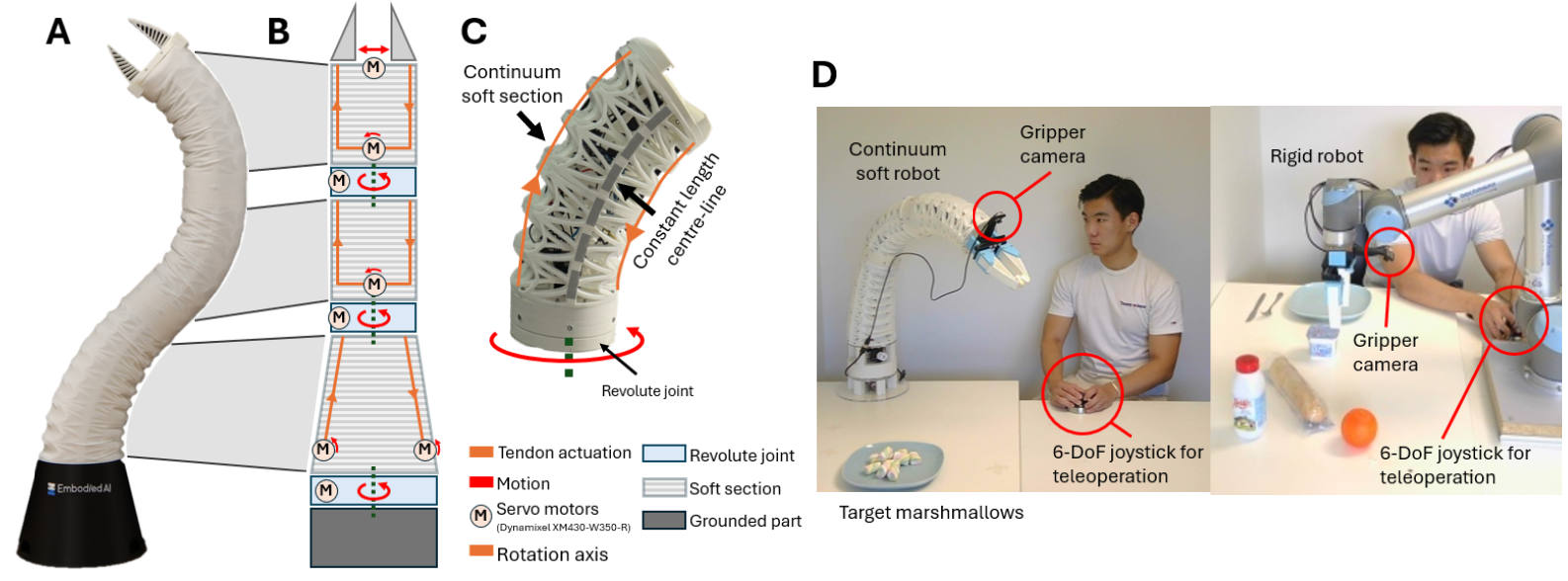}
    \caption{Continuum soft robot used for the experimental study. A: The full view of the continuum robot - Embuddy. B: Actuation and structural schematic of Embuddy, indicating tendons, joints, and motors. C: Detailed view of a single section. D: Demonstration setup for the soft and rigid robot.}
    \label{fig:robot}
\end{figure}

As a benchmark, the UR5 robot was used to perform manipulation tasks. A parallel gripper was mounted with a monocular camera mounted above for autonomy (see Fig.\ref{fig:robot}D right).

For the soft counterpart (Fig.\ref{fig:robot}D left), a custom designed continuum robot arm: Embuddy, is used shown in Fig.\ref{fig:robot}A. Embuddy consists of three modular sections, comprised of a standard revolute joint followed by a soft continuum segment (see Fig.\ref{fig:robot}B). The continuum segments (shown in detail in Fig.\ref{fig:robot}C) are tendon driven, and bend in one plane (constrained by an incompressible centerline). The continuum structure is fabricated through 3D printed Thermoplastic Polyurethane (TPU). 
Two key features of Embuddy allows for inherently safe interactions. Firstly, the underactuated sections mean regardless of the motor positions the sections are always deformable to external forces. Secondly the arm is lightweight (total 5kg), limiting its inertial forces.
% Each section can bend in two different directions and can rotate with respect to the others along the central axis that goes through the robot's body. 

% This one, fabricated from a flexible material as a unified body, has been shown to bring unprecedented levels of adaptability and passive safety. It allows for fluid deformations and robust operation even in unstructured, human-centric environments.

Although Embuddy follows a similar scale to a standard serial-link manipulator, with a height of 1m, its workspace is limited to the bending angle of each soft section, whereby the first section can bend 80° and the second and the third up to 50° each. In our experiments, we use the same camera setup and gripper for both the UR5 and Embuddy, ensuring a fair comparison across embodiments.

\subsection{Designed tasks}
We selected two pick-and-place tasks and one close human-interactive task for the experiments. For simplicity, we denote them as task 1, 2 and 3.
\begin{itemize}
    \item {\textbf{Task 1:}} "Put the orange in the plate" --> Simple pick-and-place
    \item {\textbf{Task 2:}} "Put the X in the plate"(X can be orange or milk) --> Pick-and-place with choices
    \item {\textbf{Task 3:}} "Feed the person with marshmallow" --> Close human-interactive
\end{itemize}

\subsection{Experimental setup}
Following the practices in OpenVLA-OFT\cite{openvla-oft} and $\pi_0$\cite{pi0}, we have both 3rd-person and wrist view cameras for capturing the scene. In each task, objects are randomly placed in the workspace. For more details on how the setup is done for each task, refer to Appendix \ref{appendix:a}.

\subsection{Data capturing and processing}
To capture the dataset, we use a joystick to tele-operate the robots. To teleoperate and control the robot in cartesian space, a Piecewise Constant Curvature (PCC) model is used\cite{article_configuration} for the inverse kinematics. By approximating every section as a constant curvature, the tendon lengths can be related to a modeled shape, which is used to determine the end-effector pose. 

In each episode of each demonstration, the captured observation consists of 3rd-person image, wrist image, proprioceptive state(end-effector pose) and language instruction(the task). As shown in Appendix \ref{appendix:a}, the captured images are cropped and scaled. Following the practice of OpenVLA-OFT\cite{openvla-oft}, we filter out the episodes when the robot has almost zero motion(for example when gripper is grabbing or releasing). Finally, we convert and pad the representations of state and action to desired ways and dimensions according to the configuration of the models. RLDS\cite{rlds} format is used for OpenVLA-OFT and LeRobot\cite{lerobot} format is used for $\pi_0$\cite{pi0}. We open source such datasets. For more details about how we do data capturing and processing for each models and tasks, refer to App. \ref{appendix:b}.

\subsection{Model finetuning and inference}
For OpenVLA-OFT\cite{openvla-oft}, due to the large number of parameters of the LLM backbone Llama 2 7B\cite{llama2}, the best practice that balances the accuracy and computational cost is to do full finetuing with the low-rank adaptation technique(LoRA)\cite{lora}. As for $\pi_0$, since the backbone VLM PaliGemma\cite{paligemma} has smaller number of parameters(3B), we conduct full finetuning. For more details, refer to App. \ref{appendix:finetuning}.

During inference, we use the same GPU that is used for finetuning for model prediction. On the local PC that is connected with the robot, we capture observations consisting of 3rd-person view image, wrist view image, proprio state and language instruction in real time. We send such observations to the remote, where the model predicts an action chunk based on the observation and sends the chunk back to local. The local executes the actions and captures observations again. We do this communication non-stop, until the task is done or it reaches maximum steps.

\section{Results}

In this section, we evaluate the performance and behavior of VLA models, more specifically OpenVLA-OFT\cite{openvla-oft} and $\pi_0$\cite{pi0} on Embuddy with our designed tasks. Following the most common evaluation method, we estimate the model prediction accuracy by success rate in 10 trials.
Our first experiment evaluates the performance of vanilla OpenVLA-OFT and $\pi_0$ on Embuddy, with particular interest in $\pi_0$, which is known for its stronger generalization capability. All out-of-the-box models without finetuning fail in our setting. As expected, the primary cause lies in the discrepancy between soft-robot and rigid-robot dynamics, specifically the mapping from end-effector pose to internal configurations. Due to the maximum bending angle constraints of each section, Embuddy consistently gets stuck mid-execution when the model generates motions suitable for rigid manipulators but incompatible with Embuddy’s kinematics. This result highlights the significant domain gap between rigid and soft robots, underscoring the necessity of finetuning for effective policy transfer.

\begin{figure}[h] 
    \centering
    % First row of images 
    \subfloat{%
        \includegraphics[width=0.45\columnwidth]{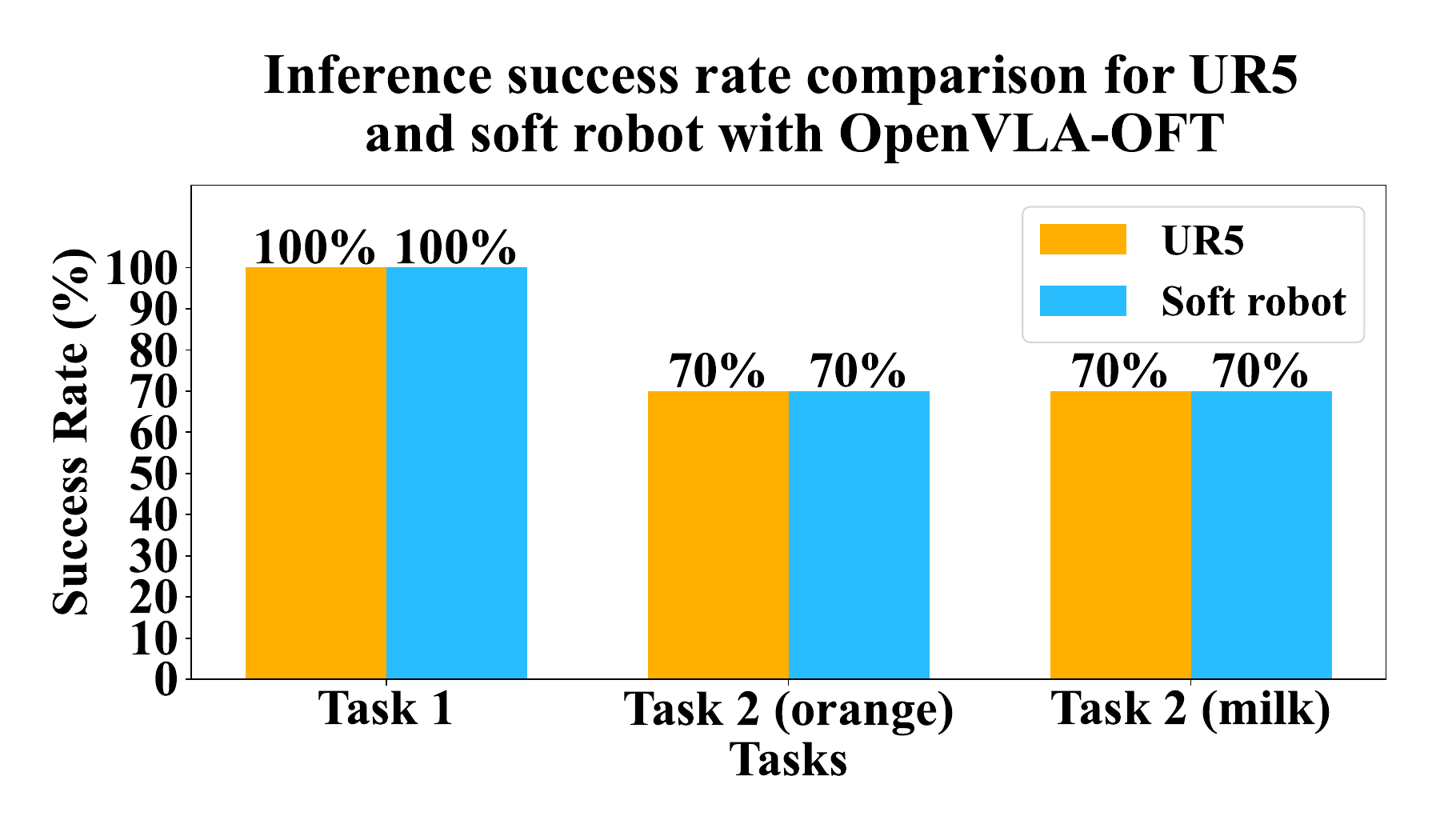}
    }
    \subfloat{%
        \includegraphics[width=0.45\columnwidth]{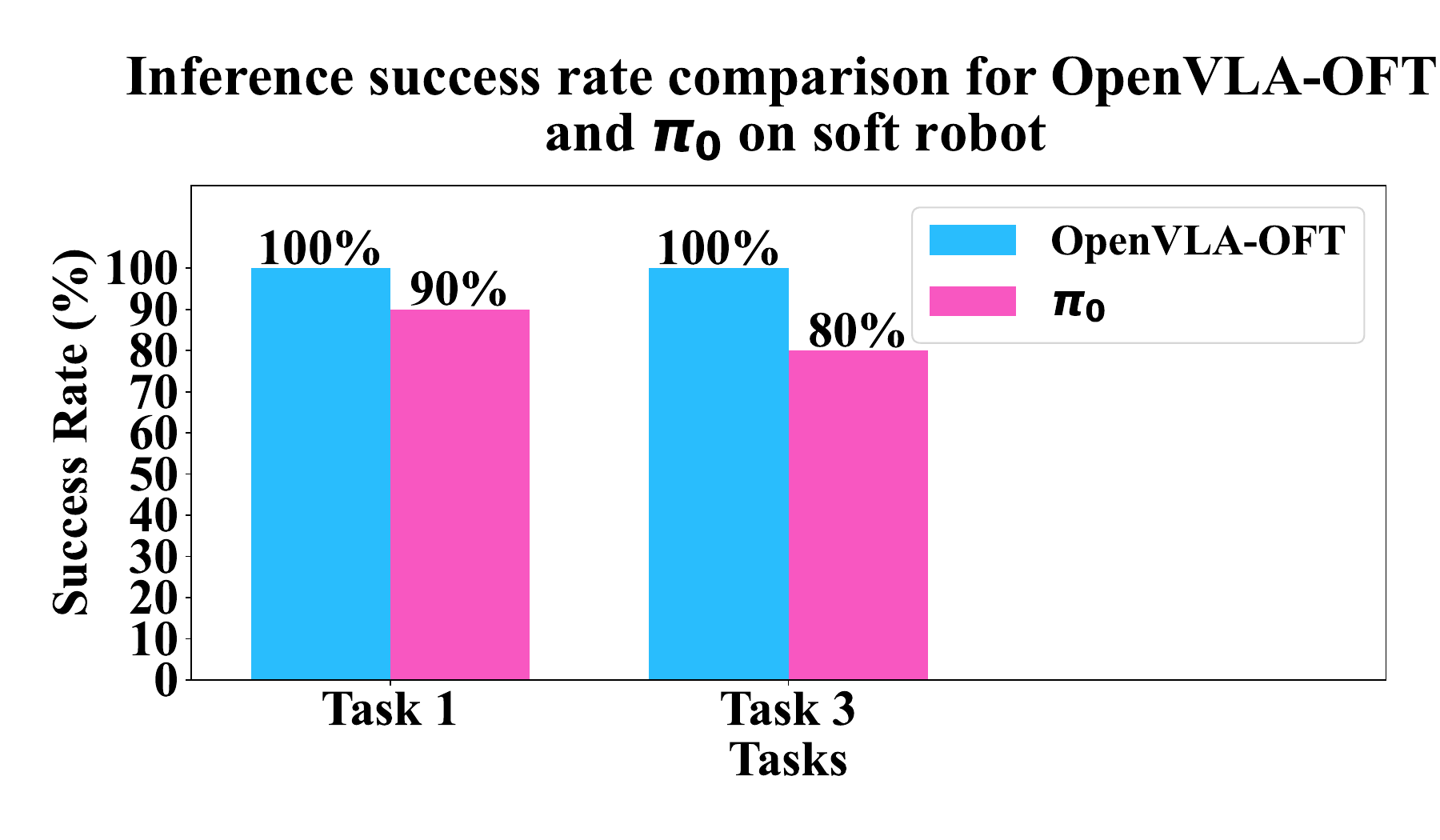}
    }

    \caption{Inference success rate comparisons between OpenVLA-OFT and $\pi_0$ on UR5 and Soft Robot embodiments.}
    \label{fig:bars}
\end{figure}

As shown in Figure \ref{fig:bars} left, applying finetuned OpenVLA-OFT\cite{openvla-oft} on UR5 and Embuddy achieve exact same success rate on task 1 and 2. This demonstrates that our finetuning strategy successfully bridges the rigid-to-soft domain gap, enabling the models to achieve comparable performance on both soft and rigid robots. Not only OpenVLA-OFT works on soft robot after finetuning, so does $\pi_0$. As shown in figure \ref{fig:bars} right, $\pi_0$ achieves high success rate, though slightly lower than OpenVLA-OFT on soft robot in task 1 and 3. It's notable that while $\pi_0$ has better generalization in rigid embodiments, OpenVLA-OFT outperforms $\pi_0$ when transferring to a completely new platform with totally different dynamics after proper finetuning. As shown in Table \ref{table-infer-freq}, even with a big connection delay, soft robot can still achieve at least 25 Hz in the control loop with OpenVLA-OFT and $\pi_0$. Appendix \ref{appendix:d} shows visualizations and more details of our experiments during inference.

\section{Conclusion}
This paper presents the first systematic deployment of Vision-Language-Action models on a soft continuum robot, directly addressing the embodiment gap between compliant and rigid manipulators. Our experiments reveal that out-of-the-box VLA policies fail due to kinematic and dynamic mismatches. However, we demonstrate that a targeted finetuning pipeline using a small, custom dataset successfully bridges this gap. The adapted policies, particularly OpenVLA-OFT, achieve high success rates on the soft robot, comparable to a rigid UR5 baseline. This work confirms that the advanced reasoning of VLA models can be effectively combined with the intrinsic safety of soft robotics. This work shows a promising direction for developing safe, adaptable, and intelligent embodied agents for human-centered environments. Future research will expand this investigation to a wider range of tasks and compliant platforms.

\newpage

\bibliographystyle{plain}
\bibliography{neurips_2025}

\newpage
\appendix

\section{Related Work}
\label{appendix:related_work}
In this section we introduce more about the two state-of-the-art methods $\pi_0$\cite{pi0} and OpenVLA-OFT\cite{openvla-oft} we apply in our work.

\subsection{$\boldsymbol{\pi_0}$}

$\pi_0$~\cite{pi0} is a Vision-Language-Action (VLA) flow model built on top of a pretrained vision-language model (VLM) backbone PaliGemma\cite{paligemma}. It supports cross-embodiment learning by training on data from multiple robotic platforms with varying kinematics and action spaces. Key architectural and training aspects include:

\begin{itemize}
    \item \textbf{Flow-matching action expert:} Rather than discretizing actions, $\pi_0$ uses conditional flow matching to predict continuous action chunks. During training, noisy action sequences are generated, and the model learns to predict the "denoising" flow that maps noise back to true actions.
    \item \textbf{Cross-embodiment generality:} $\pi_0$ is pretrained on diverse datasets comprising seven distinct robot configurations (e.g., single-arm, dual-arm, mobile manipulators) and over 68 manipulation tasks. This enables zero-shot control across different rigid platforms.
    \item \textbf{Action chunking for temporally extended tasks:} At inference, the model outputs sequences of actions via flow trajectories, allowing temporally coherent planning and execution of complex, extended tasks.
    \item \textbf{Pretraining and fine-tuning recipe:} $\pi_0$ adopts a two-stage training paradigm: broad pretraining on large-scale diverse robot data followed by task-specific fine-tuning, analogous to modern language-model training practices.
\end{itemize}

Together, these design choices enable $\pi_0$ to perform complex robotic manipulation tasks—such as laundry folding, object assembly, and mobile manipulation—via both direct prompting and fine-tuning, achieving strong generalization across embodiments and task domains.

\subsection{OpenVLA-OFT}
OpenVLA-OFT~\cite{openvla-oft} is a recent state-of-the-art Vision-Language-Action model designed to improve both performance and inference efficiency over prior VLA systems. It builds upon the OpenVLA framework~\cite{openvla}, using a ViT-based visual encoder and Llama 2 7B\cite{llama2} as the language backbone, but introduces several key innovations:
\begin{itemize}
\item \textbf{Parallel decoding with action chunking:} Instead of autoregressive token-by-token prediction, OpenVLA-OFT maps multimodal inputs directly to a sequence of actions in a single forward pass. By conditioning on empty action embeddings of length $K$ with bidirectional attention, the model predicts $K$ consecutive actions simultaneously, enabling fast execution without intermediate replanning.
\item \textbf{Continuous action outputs:} Unlike the discrete tokenized actions in OpenVLA, OpenVLA-OFT directly regresses continuous control vectors. An MLP action head replaces the output embedding layer, trained via an L1 objective to match ground-truth trajectories. This design improves precision and avoids discretization artifacts.
\item \textbf{Flexible multimodal inputs:} Beyond single-view images, the model supports multi-camera observations and low-dimensional robot states. These embeddings are projected into the shared language space and concatenated for decoding, enabling richer context awareness.
\item \textbf{Language-conditioned modulation:} To strengthen grounding, OpenVLA-OFT applies FiLM~\cite{film} layers that inject task-language embeddings into visual features at each transformer block, improving instruction following in visually ambiguous settings.
\end{itemize}
These modifications allow OpenVLA-OFT to outperform prior policies such as $\pi_0$\cite{pi0} and diffusion-based RDT-1B\cite{rdt-1b} on benchmarks including LIBERO~\cite{libero} and ALOHA~\cite{aloha}, while maintaining competitive inference speed.

\subsection{Soft robot control}
A more conventional approach towards the control of soft continuum robots have been explored in the past\cite{alessi2024rod, george2018control}. A key challenge lies in modeling and perception of soft robots due to their nature of large deformation\cite{armanini2023soft}. While methods such as finite element analysis can describe its deformation, for real-time control, simplified mathematical models such as piecewise constant curvature (PCC)\cite{della2020improved} or affine curvature models\cite{stella2023piecewise} have been developed. 
Through combination with proprioceptive sensing method(through tendon lengths\cite{softrobot}, strain sensing\cite{della2020data}, inertial measurement units\cite{stella2023soft}, vision\cite{chen2024vision}), such models can be updated in real time to estimate and control its cartesian position.

\section{Experimental setup}
\label{appendix:a}

\begin{figure}[!h]
    \centering
    % First row of images 
    \subfloat[3rd-person view]{%
        \includegraphics[width=0.48\columnwidth]{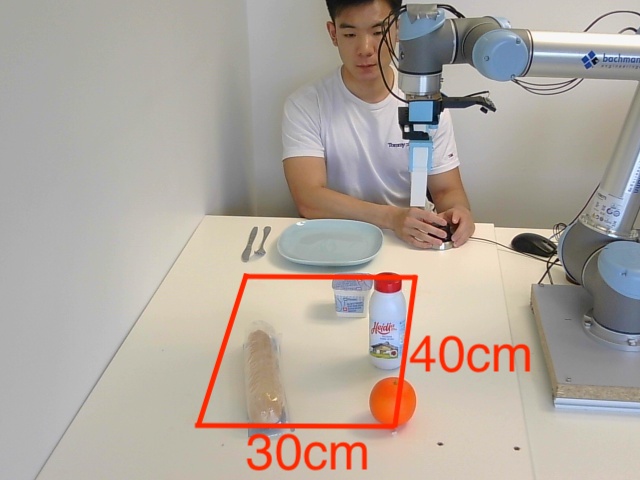}
    }\hfill
    \subfloat[Wrist view]{%
        \includegraphics[width=0.48\columnwidth]{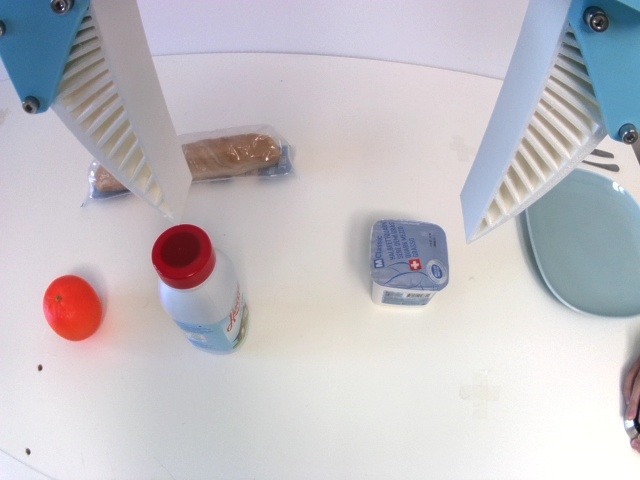}
    }

    \vspace{0.5em} % small space between rows

    % Second row of images
    \subfloat[3rd-person view after cropping and scaling]{%
        \includegraphics[width=0.48\columnwidth]{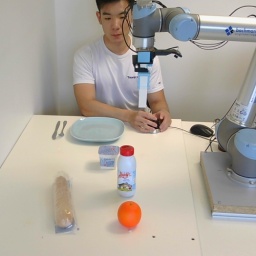}
    }\hfill
    \subfloat[Wrist view after cropping, scaling, and flipping]{%
        \includegraphics[width=0.48\columnwidth]{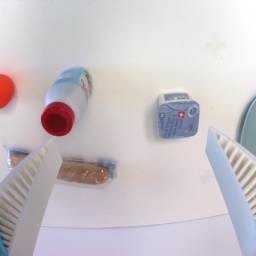}
    }

    \caption{Setup and processed image views for UR5 experiments(baseline). 
    Top row: original views, with resolution 640x480;
    Bottom row: processed views, with resolution 256x256.}
    \label{fig:ur5_setup}
\end{figure}

\begin{figure}[!h]
    \centering
    % First row of images 
    \subfloat[3rd-person view]{%
        \includegraphics[width=0.48\columnwidth]{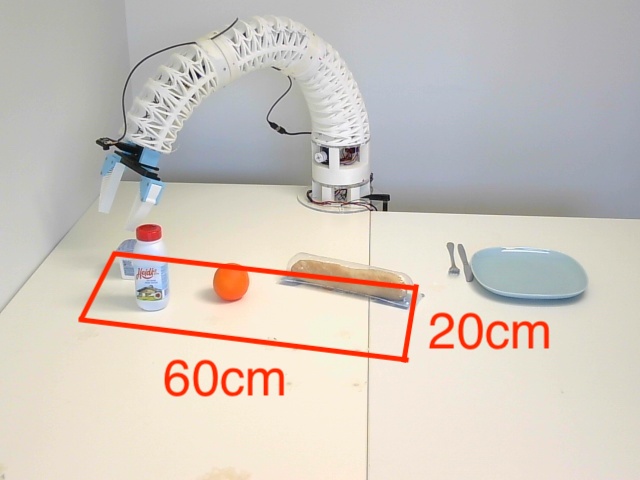}
    }\hfill
    \subfloat[Wrist view]{%
        \includegraphics[width=0.48\columnwidth]{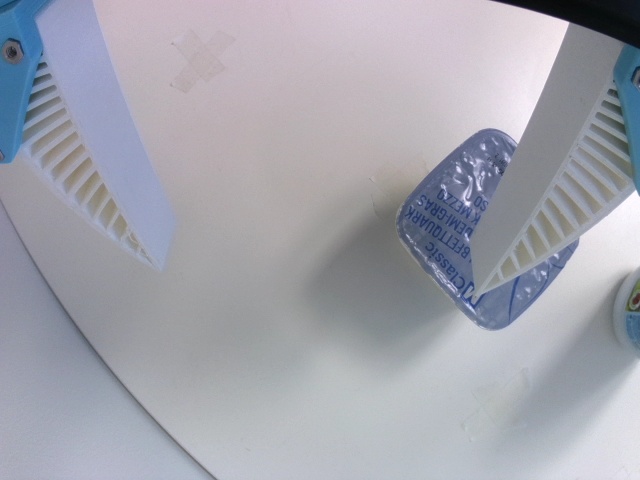}
    }

    \vspace{0.5em} % small space between rows

    % Second row of images
    \subfloat[3rd-person view after cropping and scaling]{%
        \includegraphics[width=0.48\columnwidth]{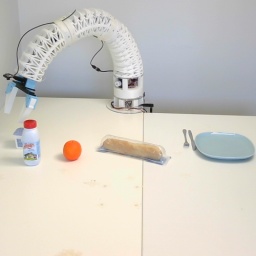}
    }\hfill
    \subfloat[Wrist view after cropping, scaling, and flipping]{%
        \includegraphics[width=0.48\columnwidth]{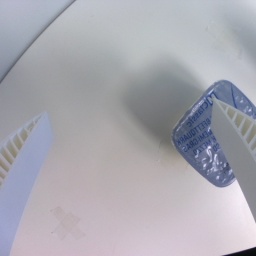}
    }

    \caption{Setup and processed image views for soft robot experiments. Top row: original views, with resolution 640x480;
    Bottom row: processed views, with resolution 256x256.}
    \label{fig:soft_setup}
\end{figure}

Here we provide extra details about our experimental setup. Figure \ref{fig:ur5_setup} shows the setup for UR5 baseline experiments. And  figure \ref{fig:soft_setup} shows the setup for soft robot experiments. Note that the workplace of two setups have same area(1200 cm$^3$), but different shapes, due to the special workspace of Embuddy. For both robots, we use the same 1 DoF gripper. And the initial end-effector pose is fixed and predefined for all tasks.

\subsection{Details for each tasks}
\textbf{Task 1: "Put the orange in the plate"} There are four common food objects(orange, milk, yogurt and baguette) that are randomly placed in the workspace. The plate is placed apart, roughly at the same place in each demonstration.

\textbf{Task 2: "Put the X in the plate"(X can be orange or milk)} Same as Task 1.

\textbf{Task 3: "Feed the person with marshmallow"} A plate of marshmallows is placed randomly in the workspace. The person in the scene stays roughly at the same position in each demonstration.

\section{Dataset capturing and processing}
\label{appendix:b}

To achieve real-time flexible 6 DoF controlling of both robots(UR5 and soft robot), we use a 3dconnexion space mouse as the joystick controller. The open/close of the gripper is controlled by the buttons on the joystick. Since we choose a relatively small gain for transformation and rotation, our capture frequency is also relatively low(5Hz). For all tasks, number of episodes in each demonstration is in the range of 50 to 200.

As shown in figure \ref{fig:ur5_setup} and \ref{fig:soft_setup}, we crop and down-sample the images to the resolution of 256 * 256. We further flip the wrist view image to make it more intuitive. 

We represent the \textbf{proprioceptive state} (pose) as an 8-dimensional vector
\[
s = \big[\, x,\, y,\, z,\, r,\, p,\, y,\, \text{pad},\, g \,\big],
\]
where $(x,y,z)$ denotes Cartesian position, $(r,p,y)$ denotes orientation in roll--pitch--yaw, 
$\text{pad}$ is a padding dimension, and $g \in \{0,1\}$ denotes the gripper state (open/closed).

The corresponding \textbf{action} is defined as the delta between adjacent poses, represented as a 7-dimensional vector
\[
a = \big[\, \Delta x,\, \Delta y,\, \Delta z,\, \Delta r,\, \Delta p,\, \Delta y,\, g \,\big],
\]
where the first six dimensions specify Cartesian and orientation increments, and 
$g \in \{0,1\}$ indicates the gripper command.

Note that roll-pitch-yaw lies within the range of $[-\pi, \pi]$. When computing the delta, directly subtracting two values near the boundaries can lead to incorrect large values. For example, the difference between $-\pi + \epsilon$ and $\pi - \epsilon$ (with small $\epsilon$) should be close to $-2\epsilon$, but a naive subtraction yields nearly $2\pi$. Therefore, the delta is handled by 
\begin{equation}
\Delta = \left( (\Delta + \pi) \bmod 2\pi \right) - \pi
\label{eq:delta}
\end{equation}

Here's the number of demonstrations captured for each task:
\begin{itemize}
    \item {Task 1: } 50
    \item {Task 2: } 100 (50 for orange; 50 for milk)
    \item {Task 3: } 20
\end{itemize}

The open source soft robot dataset can be found on HuggingFace HCSuMoss/soft\_orange and HCSuMoss/soft\_feed .

\section{Finetuning details}
\label{appendix:finetuning}

\begin{figure}[h]
    \centering
    % First row of images 
    \subfloat{%
        \includegraphics[width=0.5\columnwidth]{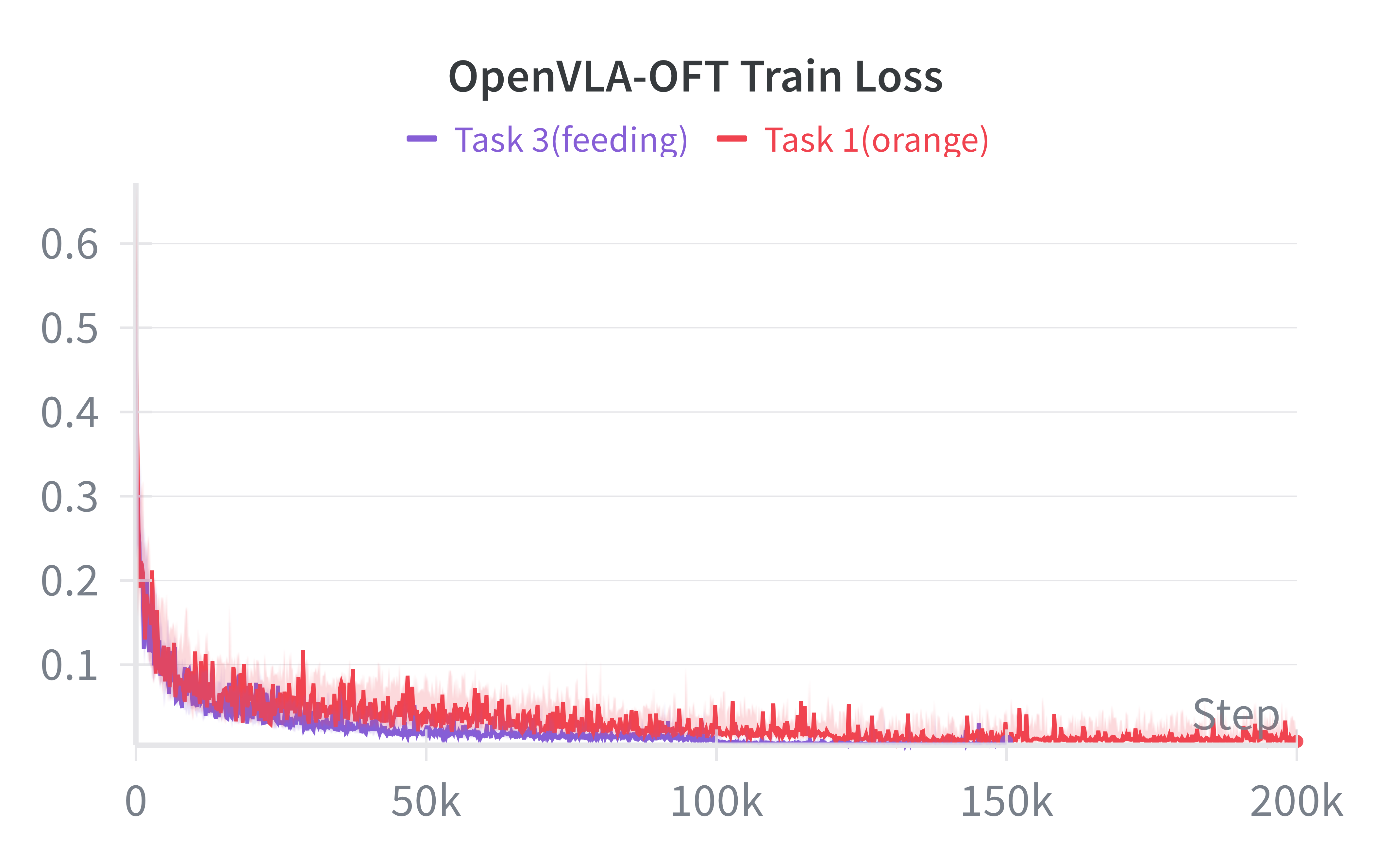}
    }
    \subfloat{%
        \includegraphics[width=0.5\columnwidth]{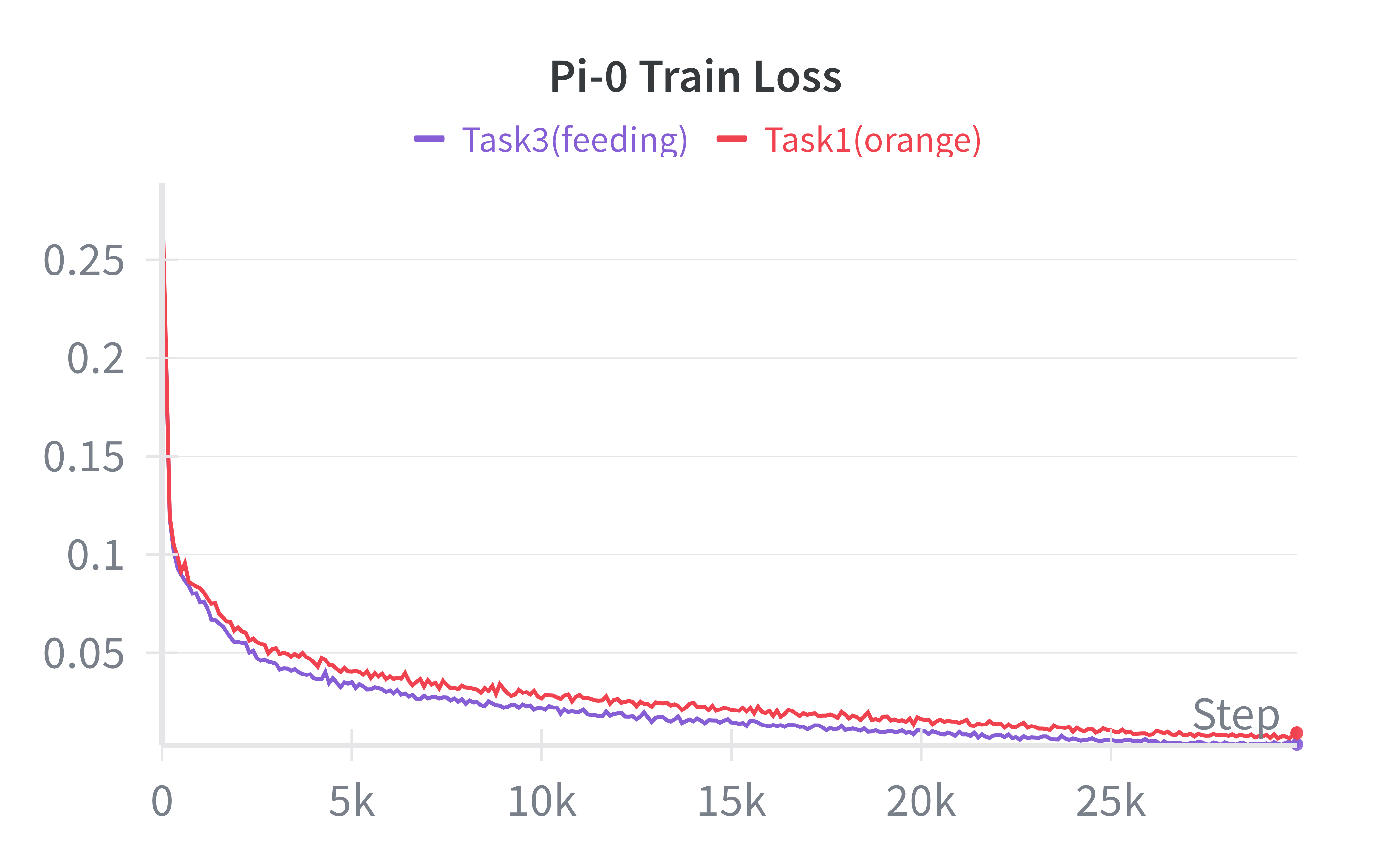}
    }

    \caption{The training losses for task 1 and 3 on soft robot with OpenVLA-OFT and $\pi_0$}
    \label{fig:loss_curves}
\end{figure}

Figure \ref{fig:loss_curves} shows the training loss curves of task 1 and 3 with OpenVLA-OFT and $\pi_0$.
\subsection{OpenVLA-OFT\cite{openvla-oft} finetuning}
According to the studies in OpenVLA\cite{openvla} paper, applying LoRA\cite{lora} with a rank of 32 is the best way to finetune on the OpenVLA-7b model, in terms of both prediction accuracy, and computational cost. To do the finetuning with such practice, we require a GPU with 80 GB(or more) memory. We utilize an A100 card on Virtual Machines of Microsoft Azure\cite{azure} cluster for UR5's experiments and an H100 card on a remote HPC cluster for soft robot's experiments.

By default data augmentation is applied on the input images, which includes augmentation with random cropping, adjustment on brightness, contrast, saturation, and hue. All the parameters are applied with default settings in the original work. 

For hyper-parameters, we mostly follow the default setting of the model. We include proprio state(pose) and two image views(3rd-person and wrist) in the input, and train continuous action head with L1 regression objective with LoRA(rank=32). The model was trained with the following hyperparameter settings:

\begin{itemize}
    \item \textbf{Action Chunk:} 8
    \item \textbf{Batch size:} 8 with one device
    \item \textbf{Learning rate:} $5 \times 10^{-4}$
    \item \textbf{Warm-up steps:} No warm-up
    \item \textbf{Learning rate decay:} After 120{,}000 steps, the learning rate decayed by a factor of 10.
    \item \textbf{Gradient accumulation:} Gradients were accumulated for 1 step, effectively applying updates at every step.
    \item \textbf{Maximum training steps:} The training process was run for a total of 200{,}000 steps.
    \item \textbf{GPU memory allocated} 63 GB
\end{itemize}

For task 2 "Put the X in the plate”, we enable the FiLM module to enhance language understanding, so that the model is capable to handle the task for both orange and milk. Due to this modification and larger dataset size, we increase the maximum training steps to 240k and adjust the learning rate decay to happen at step 180k for task 2.

And for task 3, due to a smaller amount of demonstrations included in the dataset, we reduce the maximum training steps to 150k and adjust the learning rate decay to at step 100k.

When the training loss is stabilised around 0.01, the training is done. For updating 150k steps on a single A100 card, it takes around 56 hours.

\subsection{\textbf{$\boldsymbol{\pi_0}$}\cite{pi0} finetuning}
Since the backbone of $\pi_0$ is much smaller than OpenVLA-OFT\cite{openvla-oft}, to make fair comparison, we use the full finetuning recipe for our experiments. We utilize the same H100 card on a remote HPC cluster as the experiments of OpenVLA-OFT's experiments on soft robot.

Once again, we follow the default setting and hyper-parameters of the model. To make the action chunk size same as previous experiments, we modify the action chunk size to be 8.

\begin{itemize}
    \item \textbf{Action Chunk:} 8
    \item \textbf{Batch size:} 32 with one device
    \item \textbf{Learning rate:} $2.5 \times 10^{-5}$
    \item \textbf{Warm-up steps:} 1000
    \item \textbf{Learning rate decay:} Cosine decay from warm-up to maximun training step. The final LR at the end of decay is $2.5 \times 10^{-6}$
    \item \textbf{Gradient accumulation:} Gradients were accumulated for 1 step, effectively applying updates at every step.
    \item \textbf{Maximum training steps:} The training process was run for a total of 30{,}000 steps.
    \item \textbf{Number of workers:} 2
    \item \textbf{GPU memory allocated} 91 GB (XLA\_PYTHON\_CLIENT\_MEM\_FRACTION=0.9 --> this enables JAX to use up to 90\% of the GPU memory)
\end{itemize}

For both task 1 and 3, we run the same amount of steps. It takes around 11 hours on an H100 to update 30k steps.

\section{Inference}
\label{appendix:d}
\begin{figure*}[htbp]
    \centering
    % First subfigure (top 2 rows)
    \subfloat[Inference for task 1 "put the orange in the plate" on UR5]{
        \begin{minipage}{\textwidth}
            \centering
            % Row 1: 3rd-person view
            \begin{minipage}{0.05\textwidth}
                \centering
                \small 3rd-person
            \end{minipage}
            \begin{minipage}{0.18\textwidth}
                \centering
                \includegraphics[width=\linewidth]{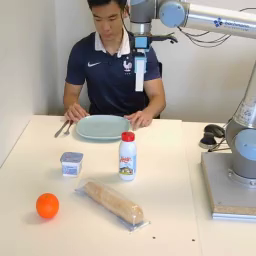}
            \end{minipage}
            \begin{minipage}{0.18\textwidth}
                \centering
                \includegraphics[width=\linewidth]{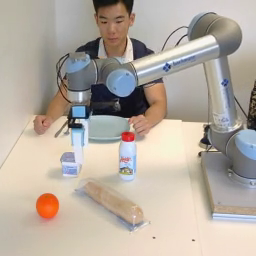}
            \end{minipage}
            \begin{minipage}{0.18\textwidth}
                \centering
                \includegraphics[width=\linewidth]{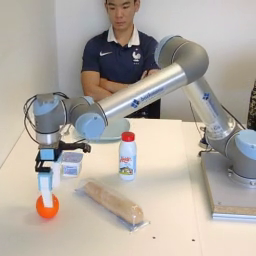}
            \end{minipage}
            \begin{minipage}{0.18\textwidth}
                \centering
                \includegraphics[width=\linewidth]{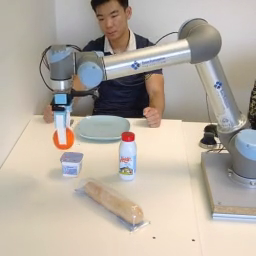}
            \end{minipage}
            \begin{minipage}{0.18\textwidth}
                \centering
                \includegraphics[width=\linewidth]{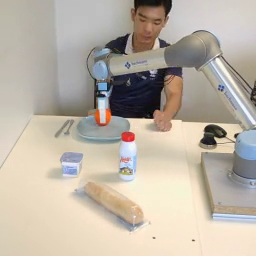}
            \end{minipage}

            \vspace{0.2em}

            % Row 2: Wrist view
            \begin{minipage}{0.05\textwidth}
                \centering
                \small Wrist
            \end{minipage}
            \begin{minipage}{0.18\textwidth}
                \centering
                \includegraphics[width=\linewidth]{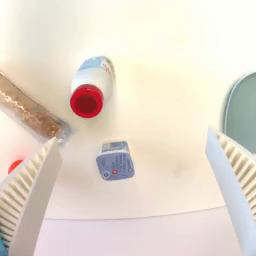}
            \end{minipage}
            \begin{minipage}{0.18\textwidth}
                \centering
                \includegraphics[width=\linewidth]{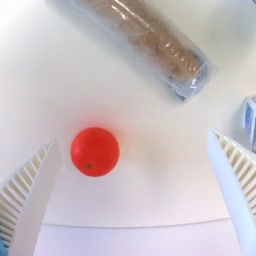}
            \end{minipage}
            \begin{minipage}{0.18\textwidth}
                \centering
                \includegraphics[width=\linewidth]{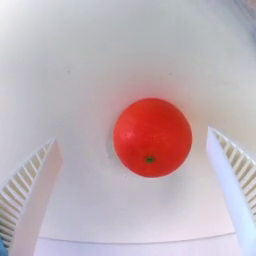}
            \end{minipage}
            \begin{minipage}{0.18\textwidth}
                \centering
                \includegraphics[width=\linewidth]{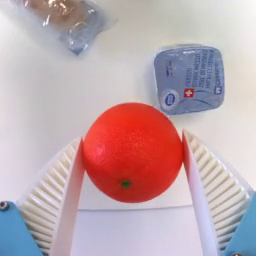}
            \end{minipage}
            \begin{minipage}{0.18\textwidth}
                \centering
                \includegraphics[width=\linewidth]{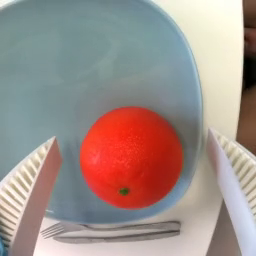}
            \end{minipage}
        \end{minipage}
        \label{fig:ur5_orange}
    }

    \vspace{0.5em}

    % Second subfigure (bottom 2 rows)
    \subfloat[Inference for task 2 "put the X in the plate" on UR5(when X is milk)]{
        \begin{minipage}{\textwidth}
            \centering
            % Row 3: 3rd-person view
            \begin{minipage}{0.05\textwidth}
                \centering
                \small 3rd-person
            \end{minipage}
            \begin{minipage}{0.18\textwidth}
                \centering
                \includegraphics[width=\linewidth]{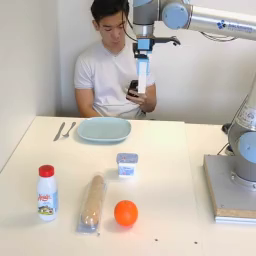}
            \end{minipage}
            \begin{minipage}{0.18\textwidth}
                \centering
                \includegraphics[width=\linewidth]{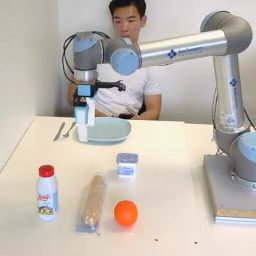}
            \end{minipage}
            \begin{minipage}{0.18\textwidth}
                \centering
                \includegraphics[width=\linewidth]{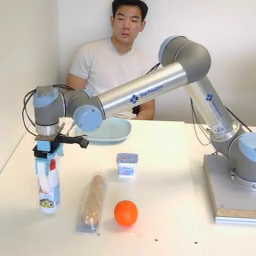}
            \end{minipage}
            \begin{minipage}{0.18\textwidth}
                \centering
                \includegraphics[width=\linewidth]{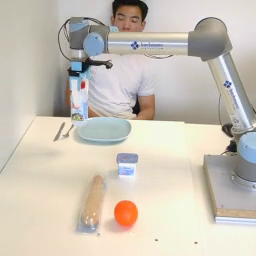}
            \end{minipage}
            \begin{minipage}{0.18\textwidth}
                \centering
                \includegraphics[width=\linewidth]{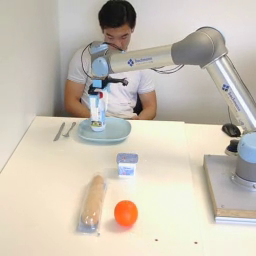}
            \end{minipage}

            \vspace{0.2em}

            % Row 4: Wrist view
            \begin{minipage}{0.05\textwidth}
                \centering
                \small Wrist
            \end{minipage}
            \begin{minipage}{0.18\textwidth}
                \centering
                \includegraphics[width=\linewidth]{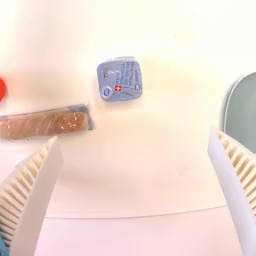}
            \end{minipage}
            \begin{minipage}{0.18\textwidth}
                \centering
                \includegraphics[width=\linewidth]{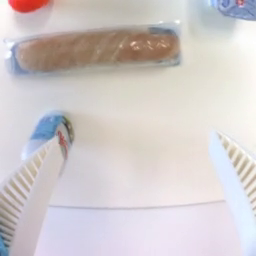}
            \end{minipage}
            \begin{minipage}{0.18\textwidth}
                \centering
                \includegraphics[width=\linewidth]{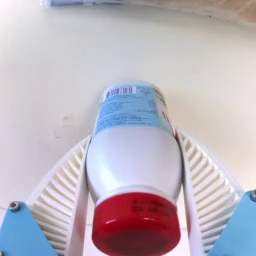}
            \end{minipage}
            \begin{minipage}{0.18\textwidth}
                \centering
                \includegraphics[width=\linewidth]{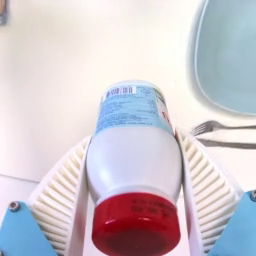}
            \end{minipage}
            \begin{minipage}{0.18\textwidth}
                \centering
                \includegraphics[width=\linewidth]{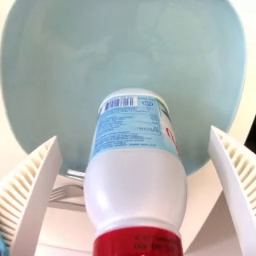}
            \end{minipage}
        \end{minipage}
        \label{fig:ur5_milk}
    }

    \caption{Visualization of the UR5 performing task 1 and 2 during inference. Each row shows 3rd-person or wrist camera views, and columns show different time steps.}
    \label{fig:ur5_inference}
\end{figure*}

\begin{figure*}[htbp]
    \centering
    % First subfigure (top 2 rows)
    \subfloat[Inference for task 1 "put the orange in the plate" on soft robot]{
        \begin{minipage}{\textwidth}
            \centering
            % Row 1: 3rd-person view
            \begin{minipage}{0.05\textwidth}
                \centering
                \small 3rd-person
            \end{minipage}
            \begin{minipage}{0.18\textwidth}
                \centering
                \includegraphics[width=\linewidth]{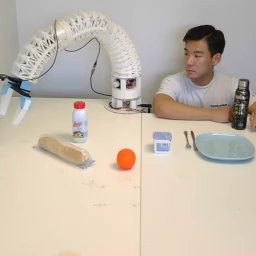}
            \end{minipage}
            \begin{minipage}{0.18\textwidth}
                \centering
                \includegraphics[width=\linewidth]{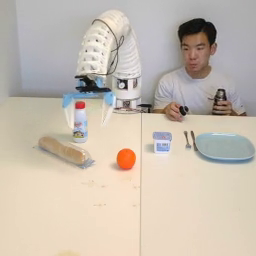}
            \end{minipage}
            \begin{minipage}{0.18\textwidth}
                \centering
                \includegraphics[width=\linewidth]{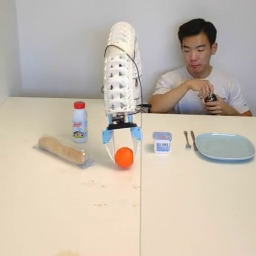}
            \end{minipage}
            \begin{minipage}{0.18\textwidth}
                \centering
                \includegraphics[width=\linewidth]{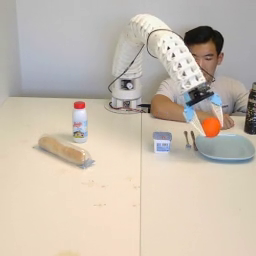}
            \end{minipage}
            \begin{minipage}{0.18\textwidth}
                \centering
                \includegraphics[width=\linewidth]{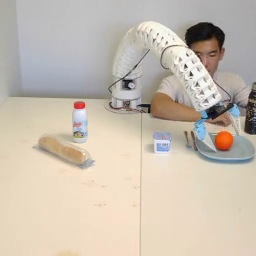}
            \end{minipage}

            \vspace{0.2em}

            % Row 2: Wrist view
            \begin{minipage}{0.05\textwidth}
                \centering
                \small Wrist
            \end{minipage}
            \begin{minipage}{0.18\textwidth}
                \centering
                \includegraphics[width=\linewidth]{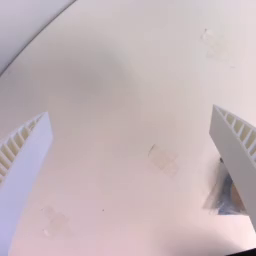}
            \end{minipage}
            \begin{minipage}{0.18\textwidth}
                \centering
                \includegraphics[width=\linewidth]{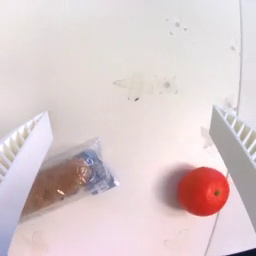}
            \end{minipage}
            \begin{minipage}{0.18\textwidth}
                \centering
                \includegraphics[width=\linewidth]{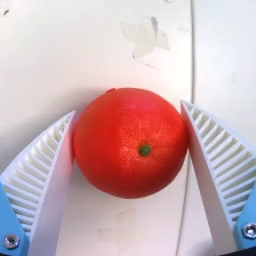}
            \end{minipage}
            \begin{minipage}{0.18\textwidth}
                \centering
                \includegraphics[width=\linewidth]{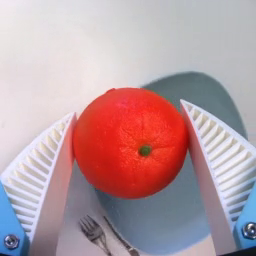}
            \end{minipage}
            \begin{minipage}{0.18\textwidth}
                \centering
                \includegraphics[width=\linewidth]{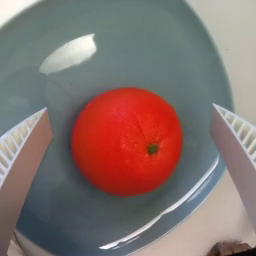}
            \end{minipage}
        \end{minipage}
        \label{fig:soft_orange}
    }

    \vspace{0.5em}

    % Second subfigure (bottom 2 rows)
    \subfloat[Inference for task 3 "feed the person with marshmallow" on soft robot]{
        \begin{minipage}{\textwidth}
            \centering
            % Row 3: 3rd-person view
            \begin{minipage}{0.05\textwidth}
                \centering
                \small 3rd-person
            \end{minipage}
            \begin{minipage}{0.18\textwidth}
                \centering
                \includegraphics[width=\linewidth]{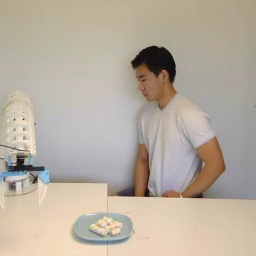}
            \end{minipage}
            \begin{minipage}{0.18\textwidth}
                \centering
                \includegraphics[width=\linewidth]{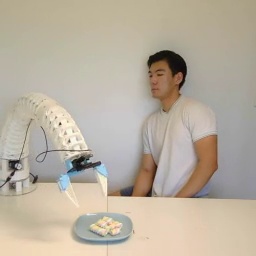}
            \end{minipage}
            \begin{minipage}{0.18\textwidth}
                \centering
                \includegraphics[width=\linewidth]{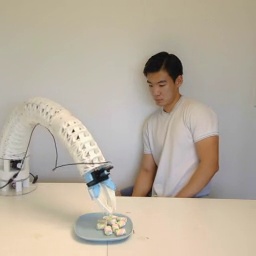}
            \end{minipage}
            \begin{minipage}{0.18\textwidth}
                \centering
                \includegraphics[width=\linewidth]{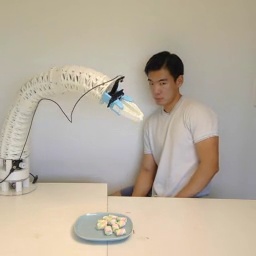}
            \end{minipage}
            \begin{minipage}{0.18\textwidth}
                \centering
                \includegraphics[width=\linewidth]{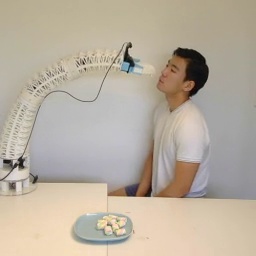}
            \end{minipage}

            \vspace{0.2em}

            % Row 4: Wrist view
            \begin{minipage}{0.05\textwidth}
                \centering
                \small Wrist
            \end{minipage}
            \begin{minipage}{0.18\textwidth}
                \centering
                \includegraphics[width=\linewidth]{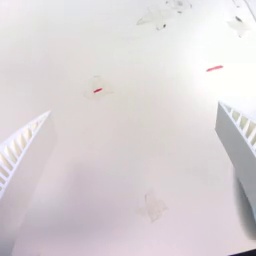}
            \end{minipage}
            \begin{minipage}{0.18\textwidth}
                \centering
                \includegraphics[width=\linewidth]{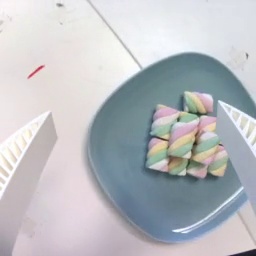}
            \end{minipage}
            \begin{minipage}{0.18\textwidth}
                \centering
                \includegraphics[width=\linewidth]{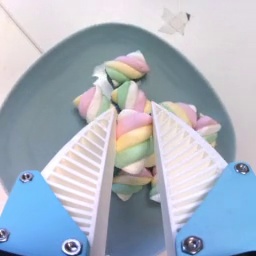}
            \end{minipage}
            \begin{minipage}{0.18\textwidth}
                \centering
                \includegraphics[width=\linewidth]{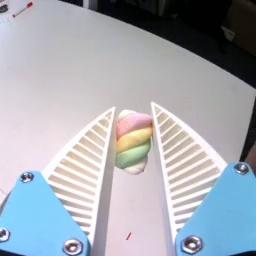}
            \end{minipage}
            \begin{minipage}{0.18\textwidth}
                \centering
                \includegraphics[width=\linewidth]{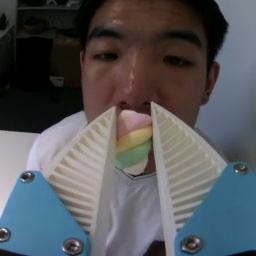}
            \end{minipage}
        \end{minipage}
        \label{fig:soft_feed}
    }

    \caption{Visualization of Embuddy performing task 1 and 3 during inference. Each row shows 3rd-person or wrist camera views, and columns show different time steps.}
    \label{fig:soft_inference}
\end{figure*}

Figure \ref{fig:ur5_inference} and \ref{fig:soft_inference} and show visualizations of our UR5 experiments and soft robot experiments on different tasks. 

\subsection{Inference perturbation}
Apart from the experiments we show in the paper, we also study about other conditions that may happen during inference in practice.

\subsubsection{With human showing in the scene}

For both experiments on UR5 and soft robot, we involve human moving freely in the scene during inference. It turns out that it has no influence on the model's performance and the model has its focus on the workspace. What's more, human can also be involved in the training set, and this is verified by both UR5 and soft robot experiments. The movement and appearance of human has zero influence in the scene as long as the workspace is not covered or interrupted. These results confirm the strong robustness of VLA models to human presence, ensuring reliable performance in human-shared environments.

\subsubsection{With unseen objects}

When there are some objects that are not shown in the training set, the model might be confused by chance. In our experiments, the model is confused once in 10 trials.

\subsubsection{When object is placed outside of workspace}

When the tasked object is placed outside of workspace, the model fails all the time, even if the object is only placed slightly(10cm) away from the region. From this test, we find that the workspace occurred in the training set is a deterministic factor, which defines the region of where the inference may succeed.

\subsection{Verification of Language Instruction}
We evaluate whether the models correctly ground their actions in the provided language instructions.

\textbf{Task 2 (Pick-and-Place with Choices):} In the “Put X in the plate” task, OpenVLA-OFT\cite{openvla-oft} achieves a 70\% success rate. The inclusion of the FiLM\cite{film} module directs the model’s attention to the object specified in the instruction, rather than selecting objects arbitrarily, indicating effective language-conditioned object selection.

\textbf{Task 3 (Human-Interactive Feeding):} In a controlled modification, we place an orange in the plate instead of marshmallows. The models(OpenVLA-OFT and $\pi_0$\cite{pi0}) appropriately refrain from executing the pick-and-place action, terminating the task mid-execution rather than incorrectly interacting with the available object. This confirms that the models' actions are semantically guided by the instruction rather than by visual salience alone.

These results collectively demonstrate that OpenVLA-OFT and $\pi_0$ reliably interprets and adheres to task-specific language instructions, supporting its robustness in language-conditioned manipulation tasks.

\subsection{Robustness of Embuddy against manual movement}
\begin{figure*}[htbp]
    \centering
    % First subfigure (top 2 rows)
    
        \begin{minipage}{\textwidth}
            \centering
            % Row 1: 3rd-person view
            \begin{minipage}{0.05\textwidth}
                \centering
                \small 3rd-person
            \end{minipage}
            \begin{minipage}{0.18\textwidth}
                \centering
                \includegraphics[width=\linewidth]{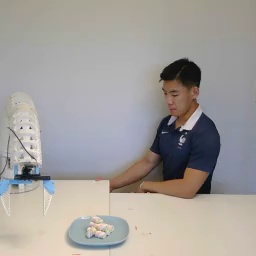}
            \end{minipage}
            \begin{minipage}{0.18\textwidth}
                \centering
                \includegraphics[width=\linewidth]{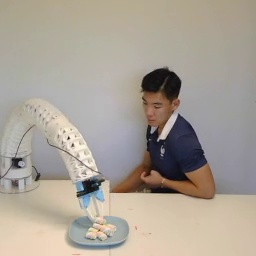}
            \end{minipage}
            \begin{minipage}{0.18\textwidth}
                \centering
                \includegraphics[width=\linewidth]{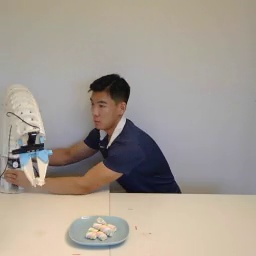}
            \end{minipage}
            \begin{minipage}{0.18\textwidth}
                \centering
                \includegraphics[width=\linewidth]{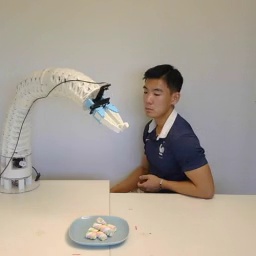}
            \end{minipage}
            \begin{minipage}{0.18\textwidth}
                \centering
                \includegraphics[width=\linewidth]{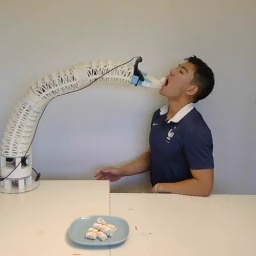}
            \end{minipage}

            \vspace{0.3em}

            % Row 2: Wrist view
            \begin{minipage}{0.05\textwidth}
                \centering
                \small Wrist
            \end{minipage}
            \begin{minipage}{0.18\textwidth}
                \centering
                \includegraphics[width=\linewidth]{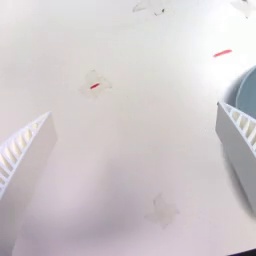}
            \end{minipage}
            \begin{minipage}{0.18\textwidth}
                \centering
                \includegraphics[width=\linewidth]{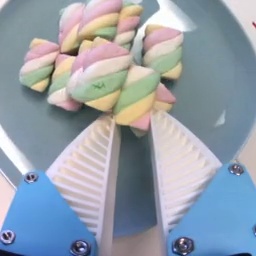}
            \end{minipage}
            \begin{minipage}{0.18\textwidth}
                \centering
                \includegraphics[width=\linewidth]{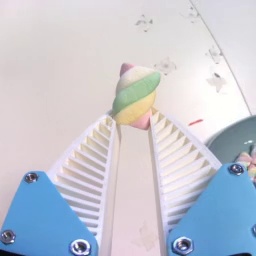}
            \end{minipage}
            \begin{minipage}{0.18\textwidth}
                \centering
                \includegraphics[width=\linewidth]{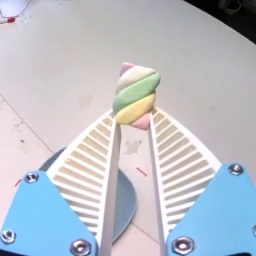}
            \end{minipage}
            \begin{minipage}{0.18\textwidth}
                \centering
                \includegraphics[width=\linewidth]{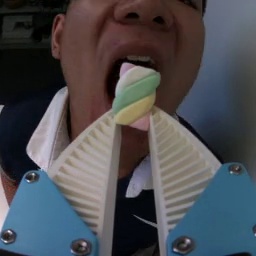}
            \end{minipage}
        \end{minipage}
        \label{fig:soft_feed_interaction}

    \caption{Visualization of Embuddy performing task 3 during inference with human interaction. As shown in the third moment, the robot's pose is manually changed by human force in the middle of inference. However, Embuddy is capable to recover it's pose and trajectory, and still complete the task successfully.}
    \label{fig:soft_inference_feed_interaction}
\end{figure*}

As one of the main advantages of soft robot in close human-interactive task is that it's bendable and easily stoppable by person. We also study the behavior of Embuddy in the VLA control loop when a person manually stops it or pushes it away. As shown in Figure \ref{fig:soft_inference_feed_interaction}, when Embuddy is manually pushed away during OpenVLA-OFT's inference stage, it can recover its original pose, continue to follow the correct trajectory and finish the task successfully without influence. In our experiments with task 3, the whole process of one trial lasts around 2 to 3 minutes. We manually stop or push away Embuddy twice, each lasts for around 5 seconds. Under such perturbation, we observe no performance degradation. 

\begin{table}[h]
  \caption{Table for inference average off-board frequency(network communication latency included). Note that within soft robot experiments, the communication latency is the same. But their latency is higher than the one in UR5 experiments. All the inference uses action chunk of size 8.}
  \label{table-infer-freq}
  \centering
  \begin{tabular}{llll}
    \toprule
    \cmidrule(r){1-2}
    Platform     & Model     & Device  & Frequency (Hz) \\
    \midrule
    UR5 & OpenVLA-OFT &A100(Azure VM) & 32.3     \\
    Embuddy     & OpenVLA-OFT &H100(Remote cluster) & 25.1      \\
    Embuddy     & $\pi_0$  &H100(Remote cluster)     & 38.0  \\
    \bottomrule
  \end{tabular}
\end{table}

\clearpage
\newpage

\end{document}